\documentclass[letterpaper, 10 pt, conference]{ieeeconf}  %

\IEEEoverridecommandlockouts                              %
\usepackage{url}

\usepackage{makeidx}         %
\usepackage{graphicx}        %
\usepackage{multicol}        %
\usepackage{amsmath}
\usepackage{amssymb}
\usepackage{subfig}
\usepackage{booktabs}
\usepackage{xspace}
\usepackage{color}
\usepackage[hang,flushmargin]{footmisc}
\usepackage{microtype}
\usepackage{float}
\usepackage{array}
\usepackage{multirow}
\usepackage{flushend}
\usepackage{cite}
\usepackage{siunitx}
\usepackage{microtype}
\DeclareMathAlphabet{\mathcal}{OMS}{cmsy}{m}{n}

\def\R{\ensuremath{\mathbb{R}}}

\def\D{\ensuremath{\mathcal{D}}}

\renewcommand{\vec}[1]{\ensuremath{\mathbf{#1}}}
\newcommand{\mat}[1]{\ensuremath{\mathbf{#1}}}
\newcommand{\norm}[1]{\left\lVert#1\right\rVert}
\newcommand{\fd}[1]{\ensuremath{\dot{#1}}}

\newcommand{\prob}[1]{\ensuremath{p\left(#1\right)}}
\newcommand{\probc}[2]{\ensuremath{\prob{#1 \;\middle\vert\; #2}}}

\newcommand{\vecM}[1]{\ensuremath{\begin{bmatrix} #1 \end{bmatrix}}}
\newcommand{\set}[1]{\ensuremath{\left\{#1\right\}}}

\DeclareMathOperator{\diag}{\text{diag}}

\DeclareMathOperator*{\argmax}{arg\,max}

\newcommand{\ie}{\mbox{i.\,e.}\xspace}

\newcommand{\etal}{\emph{et~al.}\xspace}
   
\renewcommand{\[}{\begin{equation}}
\renewcommand{\]}{\end{equation}}

\newcommand{\secref}[1]{Sec.~\ref{#1}}
\renewcommand{\eqref}[1]{Eq.~(\ref{#1})}
\newcommand{\figref}[1]{Fig.~\ref{#1}}
\newcommand{\tabref}[1]{Tab.~\ref{#1}}

\renewcommand{\baselinestretch}{0.99}

\usepackage[pdfencoding=auto, colorlinks=true]{hyperref} %

\makeatletter
\let\NAT@parse\undefined
\makeatother

\def\ourmodel{BOpt-GMM\xspace}

\begin{document}

\title{\LARGE \bf Bayesian Optimization for Sample-Efficient Policy Improvement\\in Robotic Manipulation}

\author{Adrian Röfer$^{1}$, Iman Nematollahi$^{1}$, Tim Welschehold$^{1}$, Wolfram Burgard$^{2}$, Abhinav Valada$^{1}$%
\thanks{$^{1}$ Department of Computer Science, University of Freiburg, Germany.}%
\thanks{$^{2}$ Department of Eng., University of Technology Nuremberg, Germany.}%
\thanks{This work was funded by the BrainLinks-BrainTools center of the University of Freiburg.}
}

\maketitle

\thispagestyle{empty}
\pagestyle{empty}

\begin{abstract}
Sample efficient learning of manipulation skills poses a major challenge in robotics. While recent approaches demonstrate impressive advances in the type of task that can be addressed and the sensing modalities that can be incorporated, they still require large amounts of training data. Especially with regard to learning actions on robots in the real world, this poses a major problem due to the high costs associated with both demonstrations and real-world robot interactions. 
To address this challenge, we introduce BOpt-GMM, a hybrid approach that combines imitation learning with own experience collection. We first learn a skill model as a dynamical system encoded in a Gaussian Mixture Model from a few demonstrations. We then improve this model with Bayesian optimization building on a small number of autonomous skill executions in a sparse reward setting. We demonstrate the sample efficiency of our approach on multiple complex manipulation skills in both simulations and real-world experiments. Furthermore, we make the code and pre-trained models publicly available at \url{http://bopt-gmm.cs.uni-freiburg.de}.
\end{abstract}

\section{Introduction}
\label{sec:introduction}

Efficient methods for learning new manipulation motions in a fast and reliable manner is still an open area of research in robotics. Behavioral Cloning (BC) has become the state-of-the-art technique to address this problem~\cite{osa2018algorithmic} and shows impressive results, both in the variety of skills that are trainable, as well as the types of input modalities~\cite{chisari2022correct,hartz2023treachery,honerkamp24momallm, schmalstieg2023learning,honerkamp2023n}, be it robotic proprioception, camera data, or language. Although they are more efficient than pure reinforcement learning (RL), these approaches still require many demonstrations ($100+$) to achieve high success rates.
Far more data efficient are the approaches that fit a parameterized model of the robotic skill from data. Dynamical systems fall into this category and have been shown to be able to generate physically plausible motions that provide a high level of reactivity and robustness against perturbations in the environment~\cite{schaalnonlinear,khansari2011learning,ijspeert2013dynamical,manschitz2018mixture} unlike their neural network based counterparts. These dynamical systems can be trained from a handful of demonstrations, in some cases, using certain constraints~\cite{khansari2011learning}, even using a single one. While their sample efficiency is a great strength, it is by no means a guarantee for the model's quality. Thus, it remains an open challenge to update these models given, ideally sparse, environmental feedback.

In our previous work~\cite{nematollahi2022robot}, we addressed this challenge. Given an initial dynamical system model, in our case a Gaussian Mixture Model (GMM), we trained a Soft-Actor-Critic agent, proposing updates to the dynamical system at a fixed step interval based on sensor data. We call this fusion of SAC and GMM \emph{SAC-GMM}. We demonstrated that our approach boosts the dynamical systems' performance to $80+\%$ after around $500$ episodes of autonomous exploration. While this is relatively efficient in the domain of RL, there is still the need to further reduce the samples required for model improvement.\looseness=-1

\begin{figure}[t]
	\centering
	\includegraphics[width=1\columnwidth]{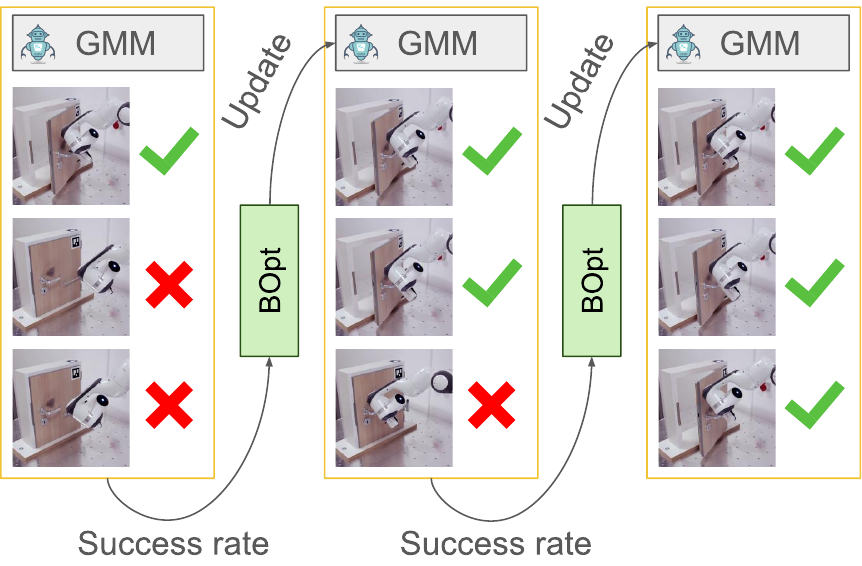}
	\caption{We propose a simple but effective interpretation of a reinforcement learning problem as black-box optimization of a policy. The policy, encoded as a GMM, is evaluated to measure its accuracy. From this new measurement, the optimizer can regress a new improved update.}
	\label{fig:introduction}
\end{figure}

If we view the problem of learning a policy not as learning a step-wise action given an observation, but rather as optimizing the value of a very costly black-box function, a perspective examined in~\cite{stulp2012policy}, we can leverage methods from the domain of black-box optimization to further improve sample efficiency. In this work, we propose \ourmodel as a fusion of sample efficient Bayesian Optimization and a GMM base policy model, as schematically represented in~\figref{fig:introduction}. 
Bayesian Optimization (BOpt) is often used for hyperparameter search in machine learning, where the evaluation of a possible set of parameters is very expensive and thus highly geared towards drawing as few samples as possible.
Previous approaches~\cite{englert2018learning,johannsmeier2019framework,wu2023prim} have used BOpt in an RL setting. Different from our approach, these works rely on predefined motion primitives with a low number of additional parameters over which they optimize, which is of very similar complexity to hyperparameter search.
In our case, we leverage BOpt to search the high-dimensional space of a multivariate GMM. The question arises as to how updates in this trajectory representation can be carried out efficiently and in a physically sound manner. 
Addressing these points, in this paper we make the following contributions:
\begin{itemize}
    \item We frame a sparse RL-setting as black-box optimization of a GMM policy model.
    \item We propose two effective, low-dimensional update methods for GMM encoded policies, which reduce the parameter space independent of the optimization scheme. We demonstrate their applicability to our optimization approach as well as a reinforcement learning baseline.
    \item We evaluate our proposed approach thoroughly in simulation as well as the real world and demonstrate a significant improvement in sample efficiency.
    \item We make the code and pre-trained models publicly available at \url{http://bopt-gmm.cs.uni-freiburg.de}.
\end{itemize}

\section{Related Work}

Learning from human demonstrations also called \emph{imitation learning}, is an approach that has been exploited for more than two decades~\cite{bain1995framework}. Its overarching goal is to learn to reproduce the actions demonstrated by a human, either externally or through teleoperation of the robot~\cite{billard2008survey,celemin2022interactive,zheng2022imitation}.
It has been deployed successfully in both robotics and autonomous driving~\cite{le2022survey}, using this approach robots have been enabled to learn many household skills~\cite{finn2017one,wong2022error,mandlekar2020learning,shridhar2023perceiver} and, lately, this technique has even been used to train large vision and language conditioned transformer models~\cite{brohan2022rt1} to enable long-horizon manipulations in LLM-based agents~\cite{ahn2022saycan}. However, training deep neural network policies from demonstrations requires hundreds or thousands of demonstrations, even when the approach is geared towards efficiency. As we are looking to deploy robots in novel environments and on novel tasks, this need for data becomes a limiting factor.

Alternatively, motions can also be learned from fewer demonstrations by encoding them in low-dimensional models. Established examples are \emph{dynamic motion primitives} (DMP)~\cite{ijspeert2013dynamical} and Gaussian Mixture Models (GMM)~\cite{khansari2011learning}. DMPs encode motions as a collection of attractors and repulsors, while GMMs encode a trajectory more directly as the statistical correlation of system state and its first-order derivative. Both can be used to learn from few human demonstrations as has been demonstrated in~\cite{khansari2011learning,figueroa2018physically,manschitz2018mixture,pairet2019learning,lu2021constrained}, with both methodologies being applied successfully in longer complex tasks~\cite{wang2022temporal,wu2023prim}. However, GMMs are the more common choice for low-dimensional imitation learning, due to their ability to encode more varied trajectories than DMPs. While it is still a rarity, in our previous work~\cite{nematollahi2022robot} we demonstrated how high-dimensional sensor information can be used in combination with GMMs to enable fast and efficient learning of reactive policies from very few demonstrations. However, the number of exploration episodes needed to perfect the policy is still quite high.

Using Bayesian Optimization methods for policy improvement is not a new approach.
BOpt has been used successfully in finding threshold parameters for walking gaits~\cite{calandra2016bayesian,rai2018bayesian} as well as efficient impact-compensating arm movements for balancing robots~\cite{kuindersma2011learning}.
In~\cite{chatzilygeroudis2017black} the authors demonstrate sample-efficient learning of a robotic policy in simple scenarios by learning function of the task dynamics as a Gaussian Process (GP). A similar technique is employed by~\cite{frohlich2019bayesian}, who additionally use a linear projection to reduce the number of policy parameters to. More flexible is the approach presented in~\cite{antonova2020bayesian} which uses an auto-encoding scheme to form a low-dimensional parameter space, trained on simulated trajectories. BOpt has also been used to decide sampling a simulation or a real world execution~\cite{marco2017virtual}, reducing the number of expensive real world examples needed to improve a policy. Different to our proposed method all of these methods require a dense reward signal.
Using binary signals in combination with BOpt has been studied as well, though less throughly. In~\cite{tesch2013expensive}, the authors formulate an approach to maximizing a stochastic binary reward by using a GP to model the parameters of a binomial distribution. While they do demonstrate that this model can be used to exploit such a function efficiently, we use a different approach to integrating binary feedback, as this model only lends itself to univariate scenarios.

Most closely related to the approach that we propose are~\cite{englert2018learning,johannsmeier2019framework,wu2023prim}. Englert~\etal~\cite{englert2018learning} propose using BOpt to identify a low-dimensional task mapping for a constrained optimization problem which they initialize from a single demonstration. They continue to sample possible mappings and finally use their motion problem with the uncovered mapping to perform control to solve their tasks. With their approach, they can optimize policies for manipulating articulated objects. They reward their agent with the negative forces exerted during the interactions. While their data efficiency is impressive, the space of their mapping is extremely low-dimensional with no more than three dimensions.
Johannsmeier~\etal~\cite{johannsmeier2019framework} model a manipulation skill as a chain of primitive controllers with learnable parameters for desired contact force and superimposed oscillations. They study the suitability of different optimizers for three example tasks and find BOpt to be sufficient, though it is outperformed by an evolutionary strategy. The low success of BOpt might be explainable with their 30-dimensional parameter space and rather large parameter range. In their setups, they derive a cost from the execution duration of a robotic skill.
Wu~\etal~\cite{wu2023prim} follow up the efforts of~\cite{johannsmeier2019framework,voigt2020multi} by again employing BOpt to learn parameters for motion primitives. Their key change is encoding the demonstration trajectories as a GMM and incorporating the probability of the current trajectory under this distribution in the optimizer's objective function, enforcing a similarity of generated trajectories to the original demonstrations. This similarity function is the main reward signal for the optimizer, with only minor influence given to task success. Otto, Celik, ~\etal~\cite{otto2023deep,celik2022specializing} phrase a policy improvement problem on DMPs similar to ours, but use a deep function approximator which leads to a significant data requirement.

Our approach differs from the discussed works in three main points: 1) We do not assume the existence of predefined control primitives or motion models but learn these fully as reactive systems from demonstration data. 2) We optimize directly over our policy model which is a much larger parameter space. 3) We use a simple binary reward signal to guide our optimization and do not rely on measures that require prior knowledge of the task, such as trajectory length, or interaction forces.
To the best of our knowledge, we are the first to use BOpt to optimize a GMM-encoded policy.

\section{Problem Formulation}
\label{sec:problem}

In this work, we consider a sparse reinforcement learning setting, in which a policy $\pi_\theta$ processing observations $s_t$, produces an action $a_t$, and receives a reward $r_t \in \R$ in return. The objective of the policy is to accumulate the maximum possible reward $R$ for an episode. We assume the reward to be sparse, only given out at the end of an episode for either success or failure, \ie $R \in \set{0, 1}$.
We assume the policy $\pi_\theta$ to be parameterized under a space $\Theta \subseteq \R^m$ and assume the existence of an update function $\oplus$ which can be used to derive an updated policy $\pi_{\theta,i} = \pi_\theta \oplus \Delta \theta_i$. We strive to find the optimal update $\Delta\theta^*$ which will yield the optimal policy $\pi^*=\pi_\theta \oplus \Delta \theta^*$. Performance is assessed by a non-deterministic evaluation function $h_{\theta}(\Delta\theta, j) \rightarrow \R$ which executes the policy yielded by the update for $j$ episodes and averages the rewards obtained by the executions. The overall objective is
\[
\Delta\theta^* = \argmax_{\Delta\theta} h_{\theta}(\Delta\theta, j),
\]
where $j$ is constant. Each evaluation of $h_\theta$ yields a data point $(\Delta\theta_i, R_i)$, which is collected in a dataset $\D = \set{(\Delta\theta_1, R_1), \ldots, (\Delta\theta_I, R_I)}$. This dataset can be used in the search for $\Delta \theta^*$. 

\begin{figure*}[t]
    \centering
    \vspace{2mm}
    \includegraphics[width=0.9\textwidth]{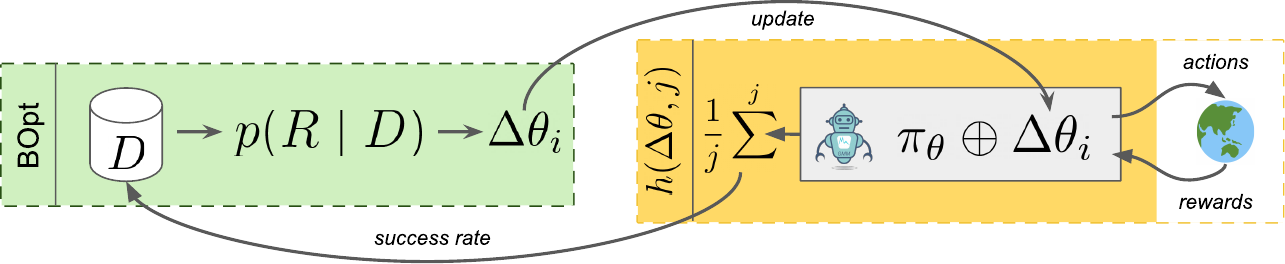}
    \caption{Our approach consists of two parts: A Bayesian optimizer estimating the value $\probc{\Delta\theta}{D}$ and proposing potential new updates $\Delta\theta_i$. The second part is the evaluation function $h(\Delta\theta_i, j)$ which plays the update $\Delta\theta$ for $j$ steps and averages the returns. The results are used to inform the optimizer.}
    \label{fig:approach}
\end{figure*}

\section{\ourmodel Framework}
\label{sec:approach}

To address the formulated problem, we introduce our approach \ourmodel. It consists of a gradient-free Bayesian optimizer which generates updates $\Delta\theta$ for the policy $\pi_\theta$ which is a dynamical system encoded as a Gaussian Mixture Model. In the following, we describe all three of these components in detail. 

\subsection{GMM}

In this work, we examine the challenge of improving an initial robotic motion policy $\pi_\theta$ trained from a set of demonstrated trajectories $\Xi_D$. Our approach assumes motion to be driven by a dynamical system of the form $f_{\theta}(s) = \fd{s}$, encoded in a GMMs as parameterization of $\pi_\theta$. Here $\fd{s}$ denotes the first-order derivative of the observable state $\vec{s}$.
We follow the Dynamical System definition of~\cite{khansari2011learning,figueroa2018physically}. Given an observable system state $\vec{s} \in \R^m$, in our case the position of the robot's endeffector relative to a frame of reference, a GMM models the dynamics of this system as $K$ components, weighted by $\vec{\omega} \in [0, 1]^K$ with $\norm{\vec{\omega}} = 1$, each of which consists of a $2m$-dimensional mean $\mu^k$, and a corresponding covariance matrix $\Sigma^k$, with
\[
 \omega \in [0, 1]^K,\hspace{4mm}  \mu^k = \vecM{\mu^k_s \\ \mu^k_{\fd{s}}}, \hspace{4mm} \Sigma^k = \vecM{\Sigma^k_{s} & \Sigma^k_{s\fd{s}} \\
                                                                                      \Sigma^k_{\fd{s}s} & \Sigma^k_{\fd{s}}}.
\]

At inference time, we use Gaussian Mixture Regression (GMR) to infer $\dot{s}_t$ from $s_t$ as 
\[
    \dot{\vec{s}}_t = f_{\theta}(s_t) = \sum_{k=1}^K h^k(s_t)(\mat{A}^k s_t + b^k)
\]
with
\[
\label{eq:gmr_detail}
\begin{aligned}
    \mat{A}^k &= \Sigma^k_{s\dot{s}} (\Sigma^k_{s})^{-1}, & b^k &= \mu^k_{\dot{s}} - \mat{A}^k\mu^k_{s} \\
\end{aligned}
\]
and $h^k(s_t)$ being the normalized probability of $\probc{k}{s_t}$. For a more detailed understanding of the inference procedure, please refer to~\cite{khansari2011learning}.

\subsection{GMM Parameterization}

Although GMMs are low-dimensional models compared to common neural network architectures, they do still hold too many parameters to be exposed directly in $\Delta\theta$. In addition, properties such as positive-definiteness of the components' covariances need to be preserved during the update integration $\pi_\theta \oplus \Delta\theta$. Thus, we are concerned with finding a small space $\Delta\Theta$, which preserves the necessary properties of the GMM.
We propose to perform norm-preserving updates to the weights. The generated updates $\Delta\theta_{\omega,i}$ at step $i$ are added to $\omega$ and normalized, while a small minimum activation $\epsilon$ is enforced for numeric stability 
\begin{align}
\omega^k_i &= \frac{\max(\omega^k + \Delta\theta_{\omega^k,i}, \epsilon)}{\sum^K_j\max(\omega^j + \Delta\theta_{\omega^j,i}, \epsilon)}.
\end{align}
Updates for the means $\mu$ are integrated additively without any further post-processing. 

The number of parameters in the covariances $\Sigma^k$ is $(2\cdot|S|)^2$ and thereby quadratic in the number of degrees of freedom $|S|$. Even by exploiting the symmetry of covariances and the fact that only $\Sigma^k_{s}$ and $\Sigma^k_{s\fd{s}}$ but not $\Sigma^k_{\fd{s}}$ are relevant for inference, the number of parameters to estimate still is quadratic in $|S|$.
Therefore, we are interested in a view of $\Sigma^k$ which enables us to formulate a much lower-dimensional update space.
We propose two schemes for updating the covariance which is linear in $|S|$.
Our first scheme is based on eigenvalue decomposition as a natural lower-dimensional parameterization. Given the eigenvectors $\mat{Q}$ and diagonal matrix of eigenvalues $\mat{\Lambda}$, we form the updated covariance as 
\[
\Sigma_{x,i}^k = \mat{Q} \mat{\Lambda} \diag(\vec{\Delta\theta}_{\Sigma^k, i}) \mat{Q}^{-1},
\]
with $\vec{\Delta\theta}_{\Sigma^k, i} \in [1 - \sigma, 1 + \sigma]$. 

In our second update scheme, we have the optimizer produce updates $\vec{\Delta\theta}_{\Sigma^k, i} \in [-\sigma, \sigma]$ per component which encodes Euler rotations, which we integrate as
\[
\Sigma_{x,i}^k = R_{XYZ}(\vec{\Delta\theta}_{\Sigma^k, i})\Sigma^k_{x}.
\]

The general intuition behind these update rules stems from the interpretation of the covariance matrix in 3-dimensional space as an ellipsoid. In the first case, we assume that the direction of the correlation of the initial model is reasonable. By changing the eigenvalues of the decomposition, we restrict the optimization to a scaling of the ellipsoid axes. In the second case, the assumption is the opposite: we preserve the scaling and instead allow for a rotation of the axes. In both cases, the number of parameters is reduced to $|S|$. Note that these are only meaningful interpretations in task space and not in latent spaces such as joint space. 

We also consider a rank-1 vector-base update for $\Sigma_i^k$ based on a singular value decomposition and rank reduction. We find this update method to be less stable than the ones presented above, but we include a brief description of it for the interested reader in \secref{sec:apx}.

\subsection{Bayesian Optimization}

Bayesian Optimization (BOpt) is the state-of-the-art technique for hyperparameter tuning in automated machine learning tasks. In this domain, its problem is formulated as finding a vector of hyperparameters $\vec{x} \in A$ which maximizes model performance as $\max_{x \in A} f(x)$, where $A$ is typically a bounded hypercube in $\R^n$. Unlike model-free techniques such as grid search and random search, BOpt algorithms build a probabilistic surrogate function $\probc{y}{\vec{x}, D}$, where $D$ is a dataset of evaluated hyperparameter samples, as defined in \secref{sec:problem} and $\vec{x}$ is a newly generated hyperparameter sample. The function's probabilistic nature allows the optimizer to explore the parameter space according to the expected value as estimated by the surrogate~\cite{hutter2019automated,yang2020hyperparameter}. A common surrogate implementation is to use Gaussian Processes (GP), however, these lend themselves mostly to lower-dimensional parameter spaces and are costly to evaluate on large datasets~\cite{hutter2019automated}.
Random forests (RF) are a common alternative to using GPs in BOpt~\cite{hutter2019automated,yang2020hyperparameter}. By using multiple random trees as regressors and averaging their output, RFs are able to provide both an expectation and uncertainty estimate of a given $\vec{x}$, while being much more time and space-efficient than GPs. As our parameter space is quite large with up to $70$ parameters, and our data becomes plentiful over time, we use RFs in this work.

New samples are evaluated on the basis of an acquisition function that rates potential new samples $\vec{x}$ according to the surrogate function. While there are many acquisitions functions, \emph{expected improvement}~\cite{jones1998efficient} is criterion that is used most commonly~\cite{hutter2019automated}.
In the automated ML literature, it is often pointed out that a drawback of BOpt is its sequential nature which bars it from parallelization. This does not concern us, as we are interested in improving a policy on a single robot, and thus have no opportunity to parallelize.

We connect BOpt to our problem by setting $A = \Delta\Omega$ and $f = h$, as defined in \secref{sec:problem}. We include the averaging evaluation function $h$ as a measurement function that measures the accuracy of a proposed sample for $j$ episodes. While existing optimizers such as SMAC~\cite{lindauer2022smac3} provide functionality for optimizing stochastic functions, they do assume these to depend on a seed they provide. Since we cannot affect the state of the external world, we model $h$ as a deterministic function and reduce the variance by selecting a sufficiently large $j$.
We represent the process of surrogate update, sampling, and sample evaluation schematically in \figref{fig:approach}.

\section{Experimental Evaluation}
\label{sec:experiments}

\begin{figure*}[t]
    \centering
    \footnotesize
    \subfloat[]{\includegraphics[width=0.25\textwidth]{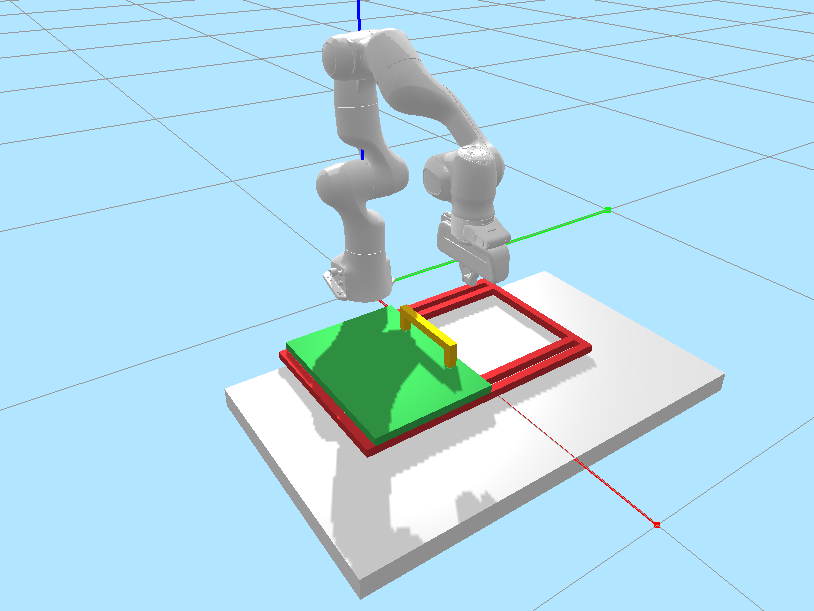}}
    \hfil
    \subfloat[]{\includegraphics[width=0.25\textwidth]{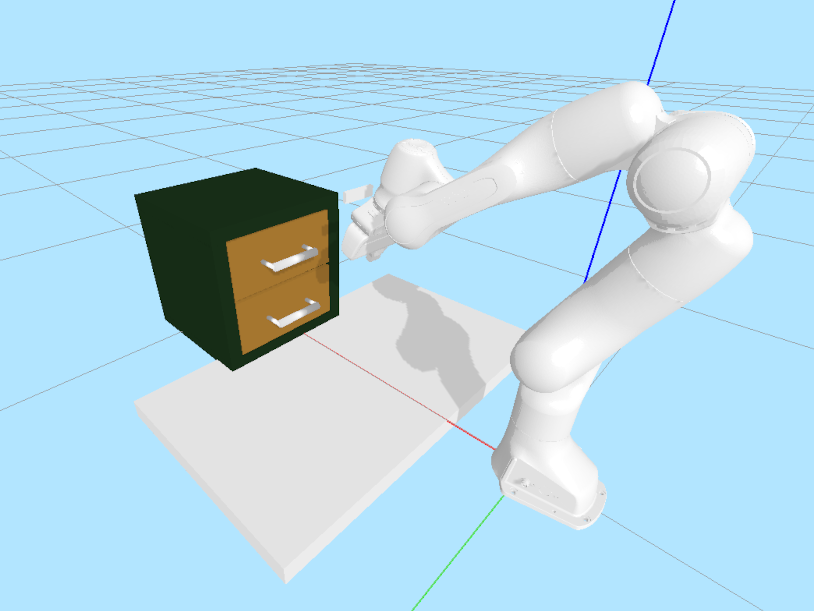}}
    \hfil
    \subfloat[]{\includegraphics[width=0.25\textwidth]{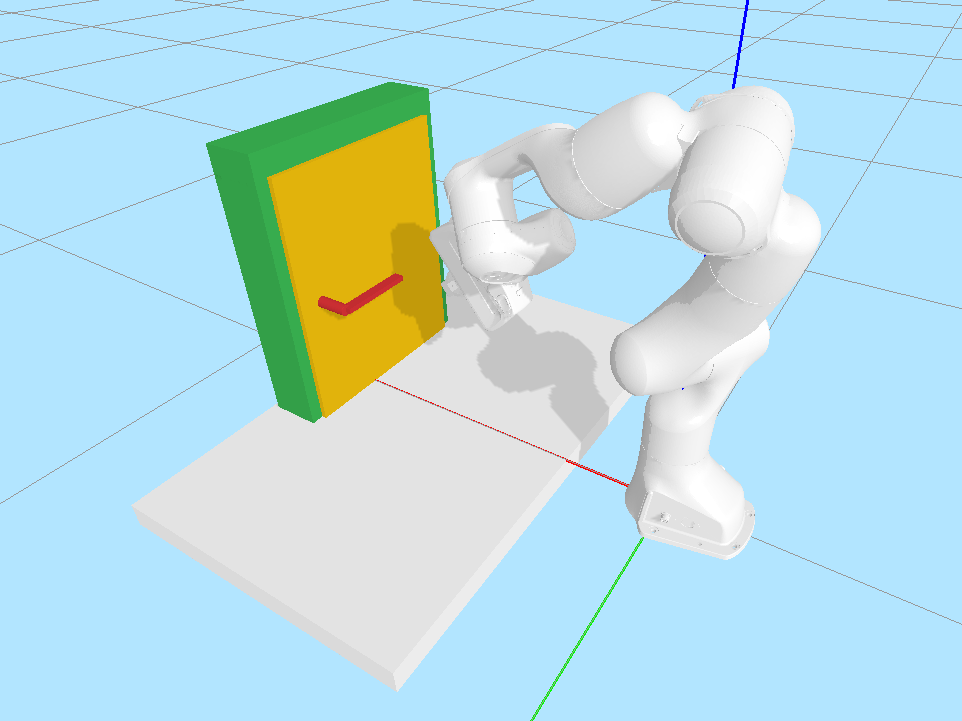}}
    \hfil\\
    \subfloat[]{\includegraphics[width=0.25\textwidth]{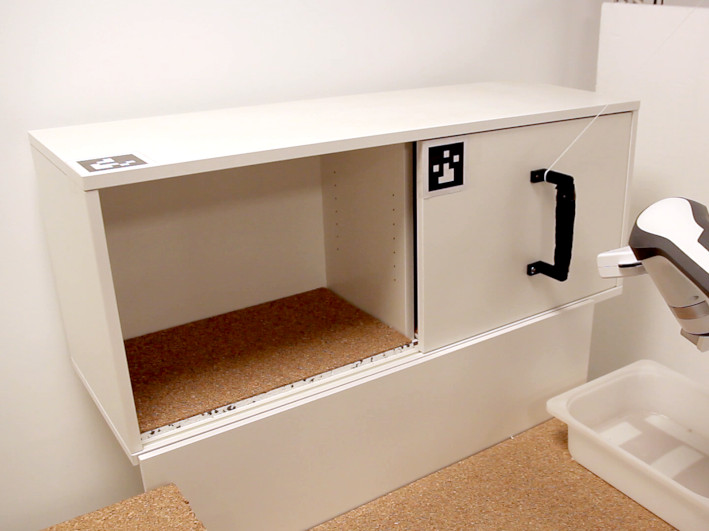}}
    \hfil
    \subfloat[]{\includegraphics[width=0.25\textwidth]{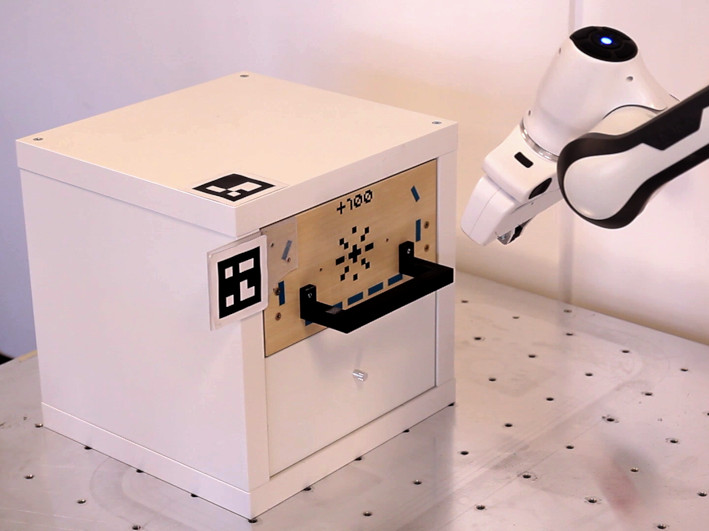}}
    \hfil
    \subfloat[]{\includegraphics[width=0.25\textwidth]{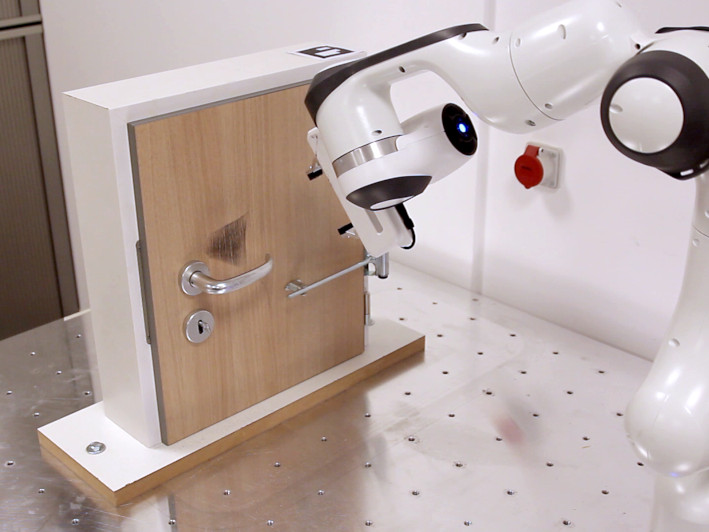}}
    \vspace{-0.2cm}
    \caption{(a) Simulated sliding of a horizontal hatch. The location and orientation of the hatch's frame are varied between episodes. (b) Simulated Drawer Opening. The location of the cabinet is varied in the XY plane. (c) Simulated opening of a door. The location of the door is varied per episode. The handle must be pressed to move the door. Real opening of a (d) sliding door, (e) drawer, (f) door.}
    \label{fig:ex_scenes}
\end{figure*}

We evaluate our proposed approach \ourmodel in 3 simulated scenarios and their matching real-world counterparts, which are shown in \figref{fig:ex_scenes}. We use the simulated scenarios not only to contrast our approach with other baselines but also to evaluate the impact of our proposed covariance update schemes.
In our real-world evaluation, we study if the well-performing optimizer configurations we have identified in simulation can also be applied to train and optimize policies directly on real robotic systems. 

\subsection{Experiment Setup}

In all evaluation scenarios, we collect 10 demonstrations by teleoperation of the robot and fit 
a GMM to the data using Expectation Maximization (EM). We also explored the usage of the SEDS framework~\cite{khansari2011learning} but did not find the resulting GMM models to improve performance while introducing additional complexity in model fitting.
Using these models as a starting point, we compare \ourmodel to two baselines:
\begin{enumerate} %
    \item[{(1)}] In simulation, we introduce a naive \emph{Online GMM} (OG) approach in which we add all successful trajectories to a growing dataset $\Xi_O$. We then refit $\pi_\theta$ to the full dataset $\Xi = \Xi_D \cup \Xi_O$ yielding our updated GMM $\pi_{\theta,i}$, where $\Xi_D$ is the set of original demonstration trajectories.
    \item[{(2)}] Our SAC-GMM approach~\cite{nematollahi2022robot} which learns an additional policy $\pi_\Delta$ which generates GMM-updates $\Delta\theta_t$ every $n$ environment steps to dynamically update the GMM. In addition to the position of the end effector, SAC-GMM receives the wrench experienced by the robot at its wrist, as we found it to learn too slowly when using solely proprioceptive observations.
\end{enumerate}

In the simulation, we also compare to a Behavior Cloning (BC) policy similar to~\cite{chisari2022correct} trained on the same initial demonstrations $\Xi_D$. As this baseline shows very limited performance, we train a variant \textit{BC 100} on an extended demonstration set $\Xi_D^{100}$.
We do not compare against plain SAC as we already determined in our previous work~\cite{nematollahi2022robot} that, due to the sparse reward setting, it does not learn any successful policy on the time horizon of interest to us. 
Since we are interested in both performance and training efficiency, we track two metrics: 1) the overall policy success rate; 2) the number of episodes taken to achieve $80\%$ success rate.

\subsection{Experiments in Simulation}

We first evaluate our approach in the simulated scenarios depicted in~\figref{fig:ex_scenes}. All of our scenarios (a-c) are manipulations of articulated objects and each poses a different challenge. The first scenario (a) requires the robot to open a sliding hatch. The robot starts above the hatch, has to loop behind the handle and push open the hatch. This task does not require greater precision but a looping steady motion.
The second scenario (b) requires the robot to open a drawer. Therefore the robot must successfully hook the handle and move in the opening direction. This task requires greater precision for the hooking of the handle but once this has been achieved it is rather forgiving. Setting (c) is the most difficult of our scenarios. Opening the door requires precise and measured motions to successfully press and hook the handle to open the door.
In all scenarios, we fit the GMMs to the relative location of the end-effector to the object. We vary the location of the object, requiring the agents to make the policy robust against variance in scenarios.

We collect 10 demonstrations in each scenario and fit a GMM for each setting. As the number $k$ of Gaussians used has an impact on the performance of the model while also scaling the number of parameters to optimize, we perform evaluations with $k \in \set{3, 5, 7}$, which yield at most $30, 50$ and $70$ parameters respectively. Throughout our experiments we use $j=8$ evaluation episodes.
We give both SAC-GMM and the Bayesian optimizer update ranges of 
\[
\begin{aligned}
\Delta\theta_\omega &\in [-0.1, 0.1] \\
\Delta\theta_\mu &\in [-0.05, 0.05] \\
\Delta\theta_\Sigma &\in [-0.1, 0.1].
\end{aligned} 
\]

\begin{figure*}[t]
    \centering
    \vspace{1mm}
    \includegraphics[width=0.95\textwidth]{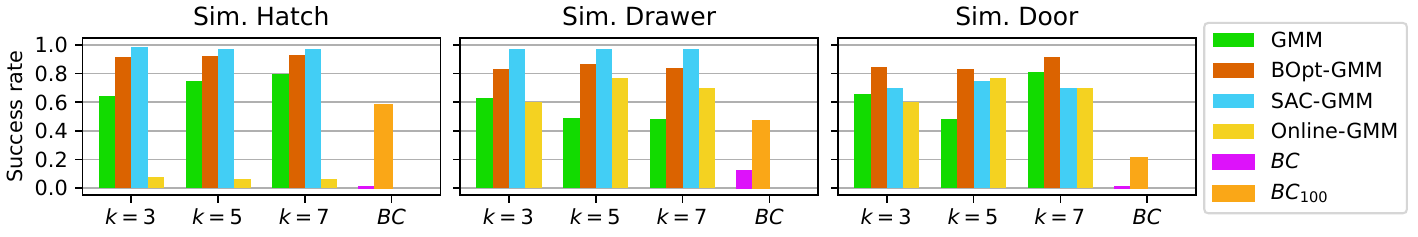}
    \caption{Comparison of the mean performances of GMM, SAC-GMM, \ourmodel, and Online-GMM baseline in our three simulated scenarios in \figref{fig:ex_scenes}). $k$ indicates the number of GMM components. We run each method for 500 episodes. We find \ourmodel and SAC-GMM to improve significantly over the initial GMM, while Online-GMM does not do so reliably, or even deteriorates performance. Additionally, we introduce $BC$ trained on the same demonstrations as the GMM, and $BC_{100}$ trained on a full $100$ demonstrations. The latter achieves recognizable but not comparable performance.}
    \label{fig:summary}
\end{figure*}

We use the Bayesian Optimization implementation from SMAC3~\cite{lindauer2022smac3}, a collection of mature gradient-free optimizers for black-box optimization.
Specifically, we use the hyperparameter optimizer with logarithmic expected improvement as an acquisition function.
In addition, we set SAC-GMM's learning rate as $2\cdot10^{-3}$. \looseness=-1
The baseline performances of these models are presented in \figref{fig:summary}.

\begin{table*}
    \centering
    \caption{Detailed analysis of the effect of GMM updates in \ourmodel and SAC-GMM. The table shows the maximal success rate of models optimized only over their means ($\mu$), their covariances using our update approaches $eig$ and $R_{XYZ}$, as well as their combinations. We optimize a GMM ($k=5$) with these update strategies using both BOpt and SAC. We report the mean final performances of our models, as well as the mean number of episodes needed to achieve $80\%$ success rate. The \emph{Mean} column reports the average of these metrics across the scenarios, while the \emph{Mean} row reports the averages across the update modalities. From the mean along both axes, we draw the overall conclusion that \ourmodel performs much faster, while SAC-GMM achieves slightly higher overall performance. From the bottom row we can also conclude that the $eig$ and $R_{XYZ}$ updates work better on their own, than when paired with $\mu$ and also work better than $\mu$ alone. We note only two exceptions to this observation in the \emph{Door} scenario for \ourmodel. }
    \begin{tabular}{ll|rr|rr|rr|rr|rr||rr}
    \toprule
     \hspace{0.4cm} &  & \multicolumn{2}{c|}{$\mu$} & \multicolumn{2}{c|}{$R_{XYZ}$} & \multicolumn{2}{c|}{$eig$} & \multicolumn{2}{c|}{$\mu + R_{XYZ}$} & \multicolumn{2}{c||}{$\mu + eig$} & \multicolumn{2}{c}{Task mean} \\
     &  &  &  &  &   &  &  &  &  &  &  &  & \\
      &  &  \multicolumn{1}{c}{Success} & \multicolumn{1}{c|}{\# eps.} & \multicolumn{1}{c}{Success} & \multicolumn{1}{c|}{\# eps.} & \multicolumn{1}{c}{Success} & \multicolumn{1}{c|}{\# eps.} & \multicolumn{1}{c}{Success} & \multicolumn{1}{c|}{\# eps.} & \multicolumn{1}{c}{Success} & \multicolumn{1}{c||}{\# eps.} & \multicolumn{1}{c}{Success} & \multicolumn{1}{c}{\# eps.} \\
     &  &  \multicolumn{1}{c}{rate} & \multicolumn{1}{c|}{$>80\%$} & \multicolumn{1}{c}{rate} & \multicolumn{1}{c|}{$>80\%$} & \multicolumn{1}{c}{rate} & \multicolumn{1}{c|}{$>80\%$} & \multicolumn{1}{c}{rate} & \multicolumn{1}{c|}{$>80\%$} & \multicolumn{1}{c}{rate} & \multicolumn{1}{c||}{$>80\%$} & \multicolumn{1}{c}{rate} & \multicolumn{1}{c}{$>80\%$} \\
     \cmidrule(l){1-14}
     \parbox[t]{2mm}{\multirow{2}{*}{\rotatebox[origin=c]{45}{Hatch}}} & 
        SAC  & $90.9\%$ & $125$ & $100.0\%$ & $75$ &  $100\%$ & $100$ & $97.5\%$ & $150$ & $98.3\%$ & $125$ & $\mathbf{97.3\%}$ & $115.0$ \\
      & BOpt & $88.8\%$ &  $28$ &  $98.6\%$ & $14$ & $99.2\%$ &  $14$ & $88.6\%$ &  $28$ & $85.6\%$ &  $28$ &          $92.2\%$ & $\mathbf{22.0}$ \\
     \cmidrule(l){2-14}
     \parbox[t]{2mm}{\multirow{2}{*}{\rotatebox[origin=c]{45}{Drawer}}} & 
        SAC  & $95.0\%$ & $150$ & $99.0\%$ & $50$ & $99.2\%$ & $50$ & $97.5\%$ & $125$ & $95.8\%$ & $225$ & $\mathbf{97.3\%}$ & $120.0$ \\
      & BOpt &   $82\%$ & $252$ & $87.6\%$ & $49$ & $94.3\%$ & $35$ & $84.6\%$ & $133$  & $88.6\%$ &  $98$ &          $87.4\%$ & $\mathbf{113.0}$ \\
     \cmidrule(l){2-14}
     \parbox[t]{2mm}{\multirow{2}{*}{\rotatebox[origin=c]{45}{Door}}} & 
        SAC  &  $79.2\%$ &   $-$ & $87.0\%$ & $250$ & $83.3\%$ & $400$ & $69.2\%$ &    $-$ & $58.3\%$ & $-$ & $75.4\%$ & $430.0$ \\
      & BOpt &    $83\%$ & $301$ & $81.6\%$ & $168$ & $85.0\%$ & $203$ & $90.3\%$ & $161$ &   $77\%$ & $-$ & $\mathbf{83.4\%}$ & $\mathbf{266.0}$\\
     \cmidrule[1.6pt](l){2-14}
     \parbox[t]{2mm}{\multirow{2}{*}{\rotatebox[origin=c]{45}{Mean}}}
      & SAC & $\mathbf{88.4\%}$ & $258.3$ & $\mathbf{95.3\%}$ & $125.0$ & $\mathbf{94.2\%}$ & $183.3$ & $\mathbf{88.1\%}$ & $258.3$ & $\mathbf{84.1\%}$ & $283.3$ & $\mathbf{90.0\%}$ & $221.6$ \\
      & BOpt & $84.6\%$ & $\mathbf{193.6}$ & $89.3\%$ & $\mathbf{77}$ & $92.8\%$ & $\mathbf{84.0}$ & $87.8\%$ & $\mathbf{107.3}$ & $83.7\%$ & $\mathbf{208.7}$ & $87.6\%$ & $\mathbf{134.1}$ \\
     \bottomrule
    \end{tabular}
    \label{tab:results_sim}
\end{table*}

\begin{figure}
    \centering
    \includegraphics[width=0.70\columnwidth]{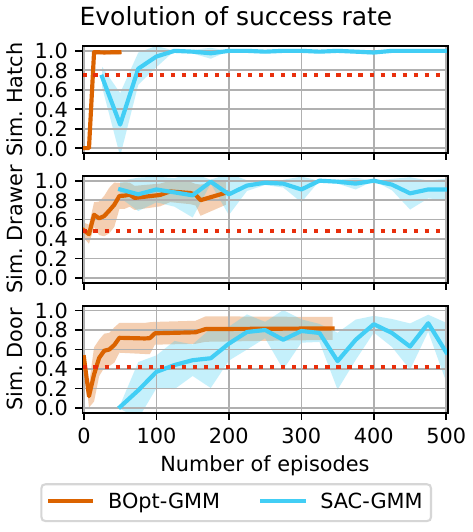}
    \caption{To illustrate the significance of the difference in sampling efficiency, we overlay the evolution of model performances of the three basic updates. The dashed red line shows the performance of the base GMM. Note: \ourmodel is only evaluated when a new incumbent is generated, while SAC-GMM is evaluated at regular intervals. Hence the different graph lengths.}
    \label{fig:results_efficiency}
\end{figure}

The success rates achieved by the approaches are reported in \figref{fig:summary}. We find that both \ourmodel and SAC-GMM improve over the baseline performance of the GMM, independent of the specific $k$. The simple Online-GMM baseline is not reliable. While it is able to increase its success rate over the starting GMM in some cases, in others it deteriorates performance dramatically. Behavioral Cloning cannot be initialized successfully from the $10$ episodes used to fit the GMMs. With an additional $90$ demonstrations, it does start to achieve noticeable performance, however, this is not the sample efficiency we aim to achieve.

While we can see from~\figref{fig:summary} that our approach and SAC-GMM work well with any $k$, we choose $k=5$ for a detailed analysis, as we have found that setting to be a good tradeoff between model performance and optimization/learning speed. For a detailed analysis, we present \tabref{tab:results_sim} where we compare the different GMM updates in SAC-GMM and \ourmodel. We find that \ourmodel achieves $80\%$ success rate $40\%$ ($87$ episodes earlier) faster than SAC-GMM. In \figref{fig:results_efficiency}, we present a qualitative comparison of the success rates of the two approaches over the training duration.
On the other hand, we also we observe that SAC-GMM achieves overall higher performance than \ourmodel.
With respect to our newly proposed covariance update strategies, we find them to achieve a much higher success rate on average in our $500$ episode timeframe than updating only the means $\mu$ as done in previous work~\cite{nematollahi2022robot}. The combination of covariance update and means update, however, does not exceed this success rate, likely due to the larger number of parameters.
A minor trend we seem to identify in our data is that the $R_{XYZ}$ update performs better for SAC-GMM than the $eig$ update, while this is reversed for \ourmodel.
We conclude from our simulated experiments that SAC-GMM yields a higher policy success rate in the long run, but \ourmodel achieves \emph{good} performance much sooner. Further, we will deploy the $R_{XYZ}$ and $eig$ updates in our real-world scenarios, as they promise to be the most effective.\looseness=-1

Finally, we would like to note a peculiarity in working with Bayesian optimization. The optimizer we use uses the surrogate model to generate a so-called \emph{incumbent} configuration which is the assumed highest performing set of parameters. The incumbent does not change after every data sample, as a new sample does not need to reveal a new optimal set of parameters. While this does not improve sample efficiency in improving the policy, it does mean that we only have to re-evaluate the performance of our policy whenever a new incumbent is generated. As can be seen in \figref{fig:results_efficiency}, in some scenarios, such as \emph{Hatch}, the best performing incumbent is found early, while in others this can take longer. These discrete moments for policy evaluation distinguish using BOpt significantly from other learning methods. It is outside of the scope of this work, but we believe there to be a potential for terminating the learning process early based on a trade-off of the remaining uncertainty in the surrogate model and the remaining possible improvement.

\subsection{Real World Experiments}

\setlength{\tabcolsep}{4.2pt}

\begin{table}
    \centering
    \vspace{2mm}
    \caption{Results of the real-world experiments. We report the maximal success rate of models and the number of episodes required to achieve $80\%$ performance. In this experiment, we find \ourmodel to outperform SAC-GMM. However, the real experiments ended after 100 episodes, which explains the comparatively low maximum performance of SAC-GMM.}
    \begin{tabular}{ll|rr|rr||rr}
    \toprule
    &  & \multicolumn{2}{c|}{$R_{XYZ}$} & \multicolumn{2}{c||}{$eig$} & \multicolumn{2}{c}{Mean} \\
      &  &  &  &  &  &  & \\
      &  &  \multicolumn{1}{c}{Success} & \multicolumn{1}{c|}{\# eps.} & \multicolumn{1}{c}{Success} & \multicolumn{1}{c||}{\# eps.} & \multicolumn{1}{c}{Success} & \multicolumn{1}{c}{\# eps.} \\
      &  &  \multicolumn{1}{c}{rate} & \multicolumn{1}{c|}{$>80\%$} & \multicolumn{1}{c}{rate} & \multicolumn{1}{c||}{$>80\%$} & \multicolumn{1}{c}{rate} & \multicolumn{1}{c}{$>80\%$} \\
     \cmidrule(l){1-8}
     \parbox[t]{2mm}{\multirow{2}{*}{\rotatebox[origin=c]{70}{Hatch}}} & 
      SAC & $63.3\%$ & $-$ & $53.3\%$ & $-$  &  $58.3\%$ & $100.0$ \\
     & BOpt & $76.6\%$ & $-$ & $85.5\%$ & $14$ & $81.1\%$ & $57.0$ \\
     \cmidrule(l){2-8}
     \parbox[t]{2mm}{\multirow{2}{*}{\rotatebox[origin=c]{70}{Drawer}}} & 
       SAC & $86.6\%$ & $25$ & $93.3\%$ & $50$ & $90.0\%$ & $37.5$ \\
     & BOpt & $86.6\%$ & $14$ & $93.3\%$ & $14$ & $90.0\%$ & $14.0$ \\
     \cmidrule(l){2-8}
     \parbox[t]{2mm}{\multirow{2}{*}{\rotatebox[origin=c]{70}{Door}}} & 
       SAC & $93.3\%$ & $50$ & $71.1\%$ & $-$ & $82.2\%$ & $75.0$  \\
     & BOpt & $93.3\%$ & $14$ & $95.0\%$ & $7$ & $94.2\%$ & $10.5$ \\
     \cmidrule[1.6pt](l){2-8}
     \parbox[t]{2mm}{\multirow{2}{*}{\rotatebox[origin=c]{70}{Mean}}} &
       SAC &          $81.1\%$ &          $58.3$ &          $72.6\%$ & $83.3$ & $76.9\%$ & $70.8$ \\
     & BOpt & $\mathbf{85.5\%}$ & $\mathbf{42.7}$ & $\mathbf{91.3\%}$ & $\mathbf{11.7}$ & $\mathbf{88.4\%}$ & $\mathbf{27.2}$ \\
     \bottomrule
    \end{tabular}
    \label{tab:results_real}
\end{table}

\setlength{\tabcolsep}{6pt}

\setlength{\tabcolsep}{4pt}

Our real-world scenarios (d, e, f in \figref{fig:ex_scenes}) mimic the scenarios we explored in the simulation. We again have a sliding door, a drawer, and a door for the robot to manipulate. We monitor the completion of the tasks using AR markers by measuring the displacement or rotation of the marker on the moving part in the frame of a static marker. 
The sliding door is registered as open when moved $\SI{0.4}{\meter}$ to the left, the drawer is considered open at $\SI{0.2}{\meter}$, and the door is at an angle of $\SI{25}{\degree}$.
The scenarios are randomized by sampling different starting locations of the robot's end-effector. 

In the real-world scenarios, we reduce the scale of the experiments. Where we verified our results in simulation on $16$ different seeds and ran each method for $500$ episodes, here we reduced the numbers to $3$ seeds and $100$ episodes.  
We use the insights we have gained from our simulations and only study the $R_{XYZ}$ and $eig$ updates. We collect $10$ demonstrations per scenario by teleoperating the robot using a gamepad and fitting an initial GMM with $k=5$ components to these. We set $\Delta\theta_\omega$, $\Delta\theta_\Sigma$, and SAC's learning rate as before.
We find our observations from the simulation experiments to be confirmed in our real-world experiments. Both SAC-GMM and \ourmodel can use our update strategies to successfully improve the baseline policy. Once again, \ourmodel passes the $80\%$ success rate threshold sooner than SAC-GMM. The minor trend we observed in our simulated experiments is born out more starkly in our real experiments: we observe that the $R_{XYZ}$ update performs better in combination with SAC-GMM, while the $eig$ update performs better with \ourmodel. Different from our simulation experiments, \ourmodel also stays ahead of SAC-GMM in overall success rate. This is likely due to the much shorter runtime of the experiments in the real-world setting.

\section{Conclusion}
In this work, we proposed employing gradient-free Bayesian Optimization (BOpt) in a sparse reinforcement learning setting with the aim of achieving greater sample efficiency. We enabled the application of BOpt to this space by encoding our underlying policy as GMM and letting BOpt find suitable updates to this policy. We coin this combination \ourmodel.
To keep the updates low-dimensional but still enable the optimizer to access the entire GMM, we proposed two low-dimensional methods for updating the GMM's covariance.
We compared our approach to three other baselines in simulation and were able to successfully deploy the parameters we identified in three real-world scenarios. In both simulation and real world, we found that our approach is significantly more efficient, achieving a success rate of $80\%$, $40\%$ faster than our baselines. We find that our covariance updates are more effective for our approach and our SAC-GMM baseline.

For future work, we see an opportunity for combining \ourmodel and SAC-GMM. \ourmodel would yield the first drastic improvement, while SAC-GMM would be tasked with developing the local reactivity needed to achieve a final couple of percentage points of success.
As a minor improvement, we believe there is an opportunity for exploiting the variance in the surrogate function to determine the number of episodes to evaluate a sample.
As a larger improvement, we are also considering if there is an early stopping criterion that could be used to reliably end the optimization process when it stagnates. While this does not improve the overall policy success rate, it would potentially further improve sample efficiency.

\section*{APPENDIX - Low-rank Covariance Updates}
\label{sec:apx}

As additional GMM size $|S|$ update, we examine a rank-1 update scheme. The scheme is based on constructing a low-rank approximation of $\Sigma^k$, performing an update under this approximation and reintegrating the update into the full-rank matrix. We do so by extracting $\mat{U}, \vec{s}, \mat{V}$ from $\Sigma^k$ as
\[
\Sigma^k - \diag(\Sigma^k) = \mat{U} \diag(\vec{s}) \mat{V}^T,
\]
using singular value decomposition. We use this decomposition to formulate a rank-1 update by only considering the first column $\mat{U}_1, \mat{V}_1$ of $\mat{U}, \mat{V}$ and the first component $\vec{s}_1$ of $\vec{s}$. Given the $i$-th update vector $\Delta\theta_{\Sigma^k,i} \in [1 - \sigma, 1 + \sigma]$, we compute the updated matrix $\Sigma_i^k$ as
\[
    \Sigma_i^k = \diag(\Sigma^k) + \mat{U}_1 \diag(\vec{s}_1) (\mat{V}_1 \cdot \Delta\theta_{\Sigma^k,i})^T.
\]
We investigate this update type in our simulated scenarios with both \ourmodel and SAC-GMM. We find that this update type is much less reliable at achieving strong performance with either method and varying number of GMM components, see~\tabref{tab:results_rank1}. Upon inspection, we find that the rank-1 update performs well ($\geq 90\%$) in some settings, but fails in others ($\leq 30\%$). We suspect this is due to the required decomposition being executed on $\Sigma^k - \diag(\Sigma^k)$, which can become numerically unstable as $\Sigma^k$ becomes more and more diagonal. However, we need this decomposition to maintain the invertibility of $\Sigma_i^k$ as required by the GMR inference, see \eqref{eq:gmr_detail} for details. In addition, this method allows us to maintain the positive determinant of $\Sigma^k$, which is required by the function $h^k(s_t)$ also required during inference.

\begin{table}[h]
    \centering
    \vspace{2mm}
    \caption{We display the average final success rates achieved by our ablated rank-1 update method. The reported success rates are averages over all simulated scenarios and all numbers of GMM components $k$. While we see in \figref{fig:summary} that \ourmodel achieves an average success rate of over $80\%$ independent of the update type and $k$, this is not the case for the low-rank update. Similarly SAC-GMM underperforms with this update. We show the aggregated performance of the other update types in the right-most column.}
    \begin{tabular}{l|r|r||r|r}
    \toprule
    & $\text{Rank-1}$ & $\mu + \text{Rank-1}$ & Mean & Mean o. \\
    &                 &                       & $\text{Rank-1}$ & Updates \\
     \cmidrule(l){1-5}
     SAC & $73.9\%$ & $72.0\%$ & $73\%$    & $88.7\%$ \\
     BOpt & $74.1\%$ & $72.1\%$ & $73.1\%$ & $87.7\%$ \\ 
     \bottomrule
    \end{tabular}
    \label{tab:results_rank1}
\end{table}

\begin{footnotesize}
    \bibliographystyle{IEEEtran}
    \bibliography{sources.bib}
\end{footnotesize}

\end{document}